\documentclass[runningheads]{llncs}

\usepackage{eccv}

\usepackage{eccvabbrv}

\usepackage{graphicx}
\usepackage{booktabs}

\usepackage[accsupp]{axessibility}  

\usepackage[pagebackref,breaklinks,colorlinks]{hyperref}

\usepackage{orcidlink}

\def\eg{\emph{e.g.}}

\def\vs{\emph{vs. }}

\usepackage{url}

\usepackage{amsthm}
\usepackage{booktabs}
\usepackage{algorithm}
\usepackage{algorithmic}
\usepackage{array}
\usepackage{multirow}
\usepackage{graphicx}
\usepackage{xcolor,colortbl}
\usepackage{color}
\usepackage{colortbl}

\usepackage{float} 
\usepackage{wrapfig}
\usepackage{marvosym}

\newcommand{\defacto}{\textit{de facto} }

\begin{document}
	
	\title{Knowledge Transfer with Simulated Inter-Image Erasing for Weakly Supervised Semantic Segmentation} 
	
	\titlerunning{Knowledge Transfer with Simulated Inter-Image Erasing for WSSS}
	
	
	
	\author{Tao Chen\inst{1} \and
		Xiruo Jiang\inst{1} \and
		Gensheng Pei\inst{1}\and \\
		Zeren Sun\inst{1} \and
		Yucheng Wang\inst{2} \and
		Yazhou Yao\inst{1}\textsuperscript{(\Letter)}
	}
	
	\authorrunning{T.Chen et al.}
	
	\institute{Nanjing University of Science and Technology,  Nanjing 210094, China \and
		Horizon Robotics, Beijing 100089, China
	}
	
	\maketitle
	
	\begin{abstract}
		Though adversarial erasing has prevailed in weakly supervised semantic segmentation to help activate integral object regions, existing approaches still suffer from the dilemma of under-activation and over-expansion due to the difficulty in determining when to stop erasing. In this paper, we propose a \textbf{K}nowledge \textbf{T}ransfer with \textbf{S}imulated Inter-Image \textbf{E}rasing (KTSE) approach for weakly supervised semantic segmentation to alleviate the above problem. In contrast to existing erasing-based methods that remove the discriminative part for more object discovery, we propose a simulated inter-image erasing scenario to weaken the original activation by introducing extra object information. Then, object knowledge is transferred from the anchor image to the consequent less activated localization map to strengthen network localization ability. Considering the adopted bidirectional alignment will also weaken the anchor image activation if appropriate constraints are missing, we propose a self-supervised regularization module to maintain the reliable activation in discriminative regions and improve the inter-class object boundary recognition for complex images with multiple categories of objects. In addition, we resort to intra-image erasing and propose a multi-granularity alignment module to gently enlarge the object activation to boost the object knowledge transfer. Extensive experiments and ablation studies on PASCAL VOC 2012 and COCO datasets demonstrate the superiority of our proposed approach. Source codes and models are available at \url{https://github.com/NUST-Machine-Intelligence-Laboratory/KTSE}.
		
	\end{abstract}

	\section{Introduction}
	\label{sec:intro}
	With huge progress in the era of deep learning, semantic segmentation has been widely applied in fields like autonomous driving and image editing \cite{long2015fully,chen2017deeplab}. Since deep learning thus far is data-driven, the model training typically involves a large abundance of labeled images. However, collecting accurate pixel-level annotation for segmentation tasks is highly labor-intensive and time-consuming \cite{bearman2016s,dai2015boxsup,chen2021semantically}. Therefore, as a promising direction to alleviate such annotation burden, weakly supervised learning has attracted the attention of many researchers. In this paper, we focus on the weakly supervised semantic segmentation (WSSS) under the supervision of image-level labels.

	\begin{figure}[t]
		\centering
		\includegraphics[width=1.0\linewidth]{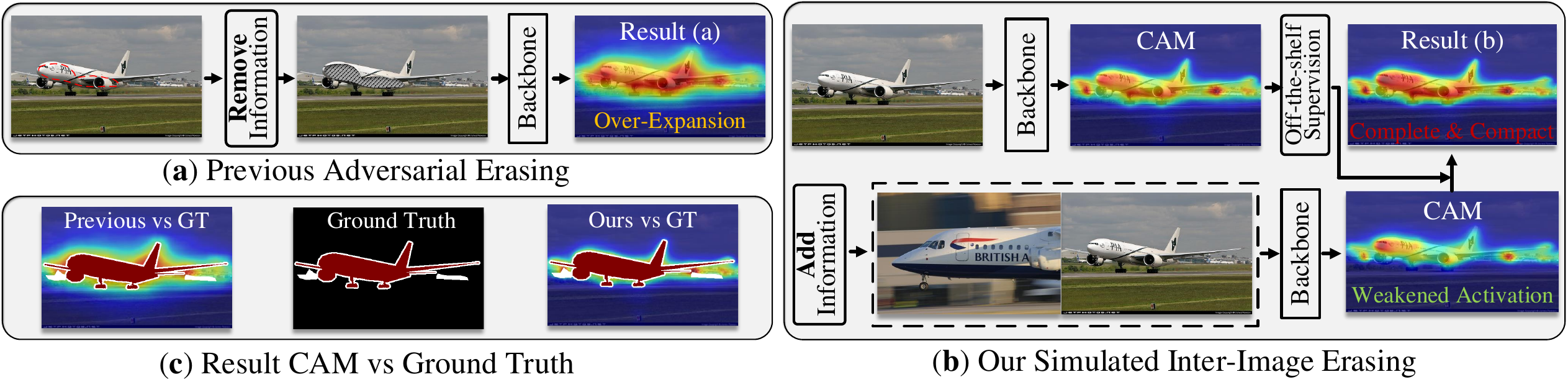}
		\caption{(a) Previous adversarial erasing-based approaches typically suffer from the over-expansion problem, which is hard to constrain. (b) Different from their information removal strategy, we propose to add extra object knowledge from a paired image to weaken the current object activation. The localization ability of the network is then enhanced by improving the consequent less activated attention map through learning from the object knowledge of the anchor branch. (C) Result comparison.}
		\label{fig_moti}
		\vspace{-0.2cm}
	\end{figure}

	The recent advance of WSSS typically follows the three-step pipeline of: 1) transforming image tags into pixel-level coarse labels, 2) refining the pseudo labels, and 3) training the final segmentation model with the refined labels. Speaking of segmentation label generation, the technique of class activation map (CAM) \cite{zhou2016learning} has been the \defacto paradigm for object localization with image-level labels. Unfortunately, the naive CAM can only highlight the most discriminative area of objects and thus its small and sparse activation will lead to incomplete object mining. Since the first step of seed generation is the foundation of later processes, numerous works have been developed to expand CAM activation for high-quality pseudo labels. Among them, adversarial erasing is one of the mainstream approaches, aiming to discover new and complementary object regions by masking the currently detected area in an adversarial manner \cite{wei2017object,zhang2018adversarial,yoon2022adversarial}. For example, the pioneering work of AE-PSL \cite{wei2017object} performs the erasing multiple times to progressively discover increasingly activated areas. The recent work of AEFT \cite{yoon2022adversarial} proposes to learn the concept of erasing with the triplet loss between the input image, erased image, and negatively sampled image. Though improvement is obtained, existing adversarial erasing-based methods still suffer from the difficulty in determining when to stop erasing: excessive removal will lead to over-expansion and insufficient erasure will result in under-activation.

	In this paper, we propose a \textbf{K}nowledge \textbf{T}ransfer with \textbf{S}imulated Inter-Image \textbf{E}rasing (KTSE) approach for WSSS to alleviate the problem mentioned above. Existing erasing-based methods typically mask the most discriminative object part to force the classifier into seeking other regions. Based on the observation that information removal leads to more activation (which might also incur the over-expansion illustrated in \cref{fig_moti} (a)), we act in a diametrically opposite way. Specifically, as shown in \cref{fig_moti} (b), we propose to simulate an inter-image erasing (SIE) scenario, introducing extra object information by concatenating a paired image and rendering the anchor image as the erased one. We then strengthen the object localization ability of the network by transferring the object knowledge from the anchor CAM to the consequent less activated localization map. Intuitively, our knowledge transfer with simulated inter-image erasing will not cause the over-expansion problem if the object in the anchor branch is well-discovered. 
	
	However, since the above knowledge transfer is bidirectional, it will also weaken object mining in the anchor branch when learning from the sparse activation of the simulated branch. Therefore, as shown in the right of \cref{fig_framework}, we propose a self-supervised regularization module for the anchor CAM feature to maintain its reliable activation in discriminative regions. Specifically, we first generate pseudo labels from the CAM feature by locating the confident foreground and background area, which are used to directly supervise the learning of CAM features. However, 
	the extracted labels of the foreground are quite noisy in complex images with multiple categories of objects due to the inter-class confusion. Therefore, we also design an inter-class loss to implicitly encourage the activation consistency in complex images to improve recognition of the inter-class object boundary. 
	
	Though our proposed self-supervised regularization can effectively constrain and facilitate the knowledge transfer in our simulated inter-image erasing module, the improvement of the network's object localization ability will be severely limited by the quality of anchor CAM feature which is usually under-activated. Therefore, we resort to the traditional intra-image erasing and propose a multi-granularity alignment (MGA) module to gently expand the object activation without introducing much background noise. Specifically, as shown in the bottom of \cref{fig_framework}, we first leverage the image-level global alignment to distill the soft object confidence from the anchor branch to the masked one for object activation enlargement. Then, we leverage a pixel-level local alignment to transfer the newly discovered object information back to the anchor branch. With the gentle and constrained activation enlargement, the proposed multi-granularity alignment module further boosts the performance of our simulated inter-image erasing. As can be seen in \cref{fig_framework}, our SIE and MGA are designed in a symmetric way that play the adversarial role of expanding and constraining the anchor image activation respectively for mining complete and compact object regions.   
	
	Our contributions can be summarized as follows:
	
	(1) We propose a knowledge transfer with simulated inter-image erasing approach for weakly supervised semantic segmentation to alleviate the over-expansion problem of existing adversarial erasing-based methods.
	
	(2) A self-supervised regularization module is proposed to constrain the knowledge transfer for maintaining reliable activation and improving recognition of the inter-class object boundary.
	
	(3) We propose a multi-granularity alignment module to gently enlarge the object activation via image-level global alignment and pixel-level local alignment, boosting the knowledge transfer with simulated inter-image erasing.
	
	(4) Extensive experiments and ablation studies on PASCAL VOC 2012 and COCO datasets demonstrate the superiority of our approach.

	\section{Related Work}
	\subsection{Weakly Supervised Semantic Segmentation}
	Weakly supervised semantic segmentation (WSSS) is the task that aims to obtain a high-performance segmentation model by learning only from weak labels. Compared with bounding boxes \cite{dai2015boxsup,khoreva2017simple,jing2019coarse,song2019box}, scribbles \cite{lin2016scribblesup,vernaza2017learning} and points \cite{bearman2016s}, the image-level label \cite{lee2021railroad,zhou2022regional,jiang2022l2g,ahn2019weakly,yao2021non,chen2022saliency,chen2024spatial,pei2024videomac,cai2024poly,sheng2024adaptive} is the cheapest weak supervision to collect, making it the most popular annotation format for WSSS.
	The recent advance of WSSS \cite{lee2021railroad,zhou2022regional,jiang2022l2g,ahn2019weakly,yao2021non} typically relies on the class activation technique of CAM \cite{zhou2016learning} to locate the target object, which helps transform the image-level label to pixel-level dense annotation for segmentation network training. Since the object area highlighted in CAM is usually small and sparse, enlarging the CAM activation becomes the focus of the WSSS task. For example, RDC \cite{wei2018revisiting} and DRS \cite{kim2021discriminative} propagate the object information to less discriminative regions via dilated convolution with varied rates and attention suppression of most discriminative areas, respectively. Popular contrastive learning \cite{du2022weakly,zhou2022regional,ru2023token} and self-supervised approaches \cite{shimoda2019self,wang2020self,chen2022self} are also applied to facilitate object mining by improving representation learning. Moreover, cross-image information \cite{fan2020cian,zhou2021group,wang2022looking} is exploited to help the classifier discover more object patterns to obtain consistent and integral target regions.
	Besides attention expansion, Ahn and Kwak propose AffinityNet \cite{ahn2018learning} for CAM refinement, which propagates semantic affinity via a random walk to recover delicate shapes of objects. IRN \cite{ahn2019weakly} is then proposed to exploit class boundary detection for discovering the entire instance areas with more accurate boundaries.

	\subsection{Erasing-based Approaches}
	Considering the original CAM technique usually leads to suboptimal performance due to failure to localize all object parts, erasing-based approaches are commonly employed to expand the highlighted region \cite{singh2018hide,zhong2020random,wei2017object,zhang2018adversarial,kweon2021unlocking,yoon2022adversarial}. They mask areas in a training image to force the network to seek other relevant parts. While direct erasing methods like Hide-and-Seek \cite{singh2018hide} hide image patches randomly, adversarial erasing-based approaches mask the most discriminative regions with the attention information, demonstrating more promising activation expansion potential. The first adversarial erasing-based work for WSSS is proposed by Wei \etal \cite{wei2017object}, which progressively discovers new object discriminative regions by erasing the currently activated area. Such a heuristic erasing strategy then attracts the attention of many researchers. For example, ACoL \cite{zhang2018adversarial} proposes adversarial complementary learning with two adversary classifiers. Recently, Kweon \etal propose to apply the adversarial erasing framework to exploit the potential of the pre-trained classifier \cite{kweon2021unlocking} and learn the concept of erasing with the triplet loss \cite{yoon2022adversarial}. While adversarial erasing approaches achieve great success in enlarging the CAM activation, they tend to suffer from the over-expansion problem due to difficulty in determining when to stop erasing. Unlike the information removal strategy adopted in these approaches, we propose a simulated inter-image erasing scenario to weaken the activation by adding extra discriminative object information. The localization ability of network is then enhanced by improving the consequent less activated CAM through knowledge transfer.

	\begin{figure*}[t]
		\centering
		\includegraphics[width=1.0\linewidth]{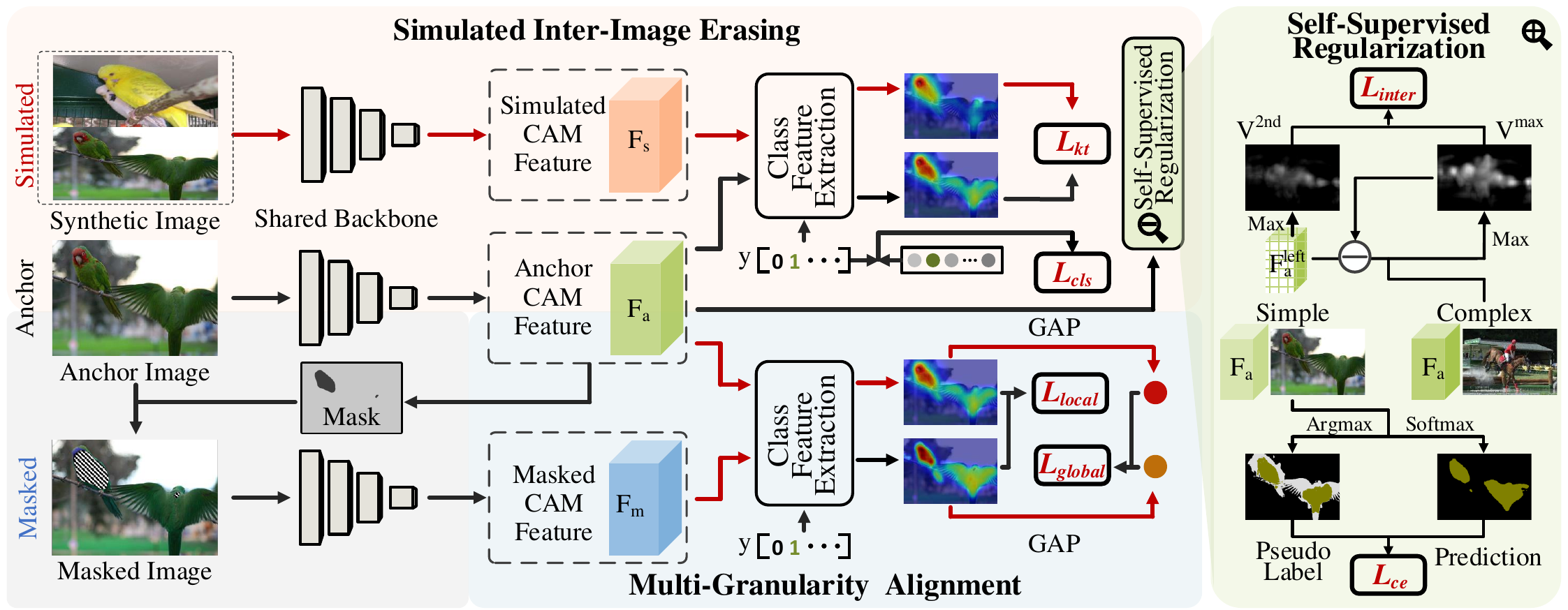}
		\caption{The architecture of our proposed approach. We propose a simulated inter-image erasing  (SIE) scenario where extra object information is introduced from another paired image. We then strengthen the object localization ability of the network by improving the consequent less activated localization map through learning object knowledge from the anchor image. A self-supervised regularization (SSR) module is also proposed to avoid weakening the anchor activation due to bidirectional alignment and improve the inter-class object boundary recognition for complex images. In addition, we propose a multi-granularity alignment (MGA) module to gently enlarge the object activation to further boost the object knowledge transfer.}
		\label{fig_framework}
		\vspace{-6pt}
		
	\end{figure*}
	
	\section{Method}
	
	In this paper, we propose a knowledge transfer with simulated inter-image erasing approach for weakly supervised semantic segmentation to alleviate the over-expansion and under-activation problems of existing adversarial erasing-based methods. Our framework is illustrated in \cref{fig_framework}. As demonstrated in the anchor branch, we train a classification network with the given image-level weak labels,  which consists of a backbone feature extractor and a pooling classification head. In contrast to existing adversarial erasing-based approaches that expand the activation through masking the most discriminative object part, as shown in the top-left of \cref{fig_framework}, we propose to simulate an inter-image erasing scenario where extra object information is introduced through concatenating a paired image. Object knowledge is then transferred from the anchor image to the consequent less activated localization map for strengthening the object localization ability of network. We also propose a self-supervised regularization module to avoid weakening the anchor activation due to bidirectional alignment and improve the inter-class object boundary recognition for complex multi-category images. In addition, we propose a multi-granularity alignment module to gently enlarge the object activation for boosting the object knowledge transfer.

	\subsection{CAM Generation}
	\label{cam_g}
	For the architecture of classification network , we follow the previous erasing-based work ACoL \cite{zhang2018adversarial} and remove the final fully-connected layer. We set the output channel of the backbone to $C+1$, where $C$ is the number of foreground categories and 1 is added for the background. We can thus directly generate object localization maps from the class-aware CAM features $F$ in the forward pass to facilitate the object knowledge transfer. To obtain CAM for each foreground class $c$, we feed the attention map $F^{c}$ into a ReLU layer and then normalize it to the range from 0 to 1:
	\begin{equation}
		A^{c}=\frac{ReLU\left ( F^{c} \right )}{\max\left ( F^{c} \right )}.
		\label{eq_norm}
	\end{equation}
	To improve the quality of the CAM by capturing the global context and local details of the image, we adopt the Gated Pyramid Pooling (GPP) layer 
	\cite{yoon2022adversarial} as the final pooling head. With the classification logits $q^{c}$ generated from the pooling head, we train the classification network with the multi-label soft margin loss as follows: 
	\begin{equation}
		\mathcal{L}_{cls}=-\frac{1}{C} \sum_{c=1}^{C} y^{c} \log \sigma\left(q^{c}\right)+\left(1-y^{c}\right) \log \left[1-\sigma\left(q^{c}\right)\right].
	\end{equation}
	Here, $\sigma\left(\cdot \right)$ is the sigmoid function. $y^{c}$ is the image-level label for the $c$-th class. 

	\subsection{Simulated Inter-Image Erasing}
	Due to the sparsity of the CAM activation, previous adversarial erasing-based methods typically enlarge the object activation by maintaining the classification confidence on masked images or features, where the most discriminative regions are erased. However, due to the absence of guidelines on when to stop erasing, these approaches easily incur over-expansion with excessive removal and might also still suffer from under-activation with insufficient erasure. Therefore, we propose a knowledge transfer with simulated inter-image erasing approach for weakly supervised semantic segmentation to alleviate the above problem.
	
	In contrast to existing erasing-based approaches that remove the discriminative part for more object mining, we propose a simulated inter-image erasing (SIE) scenario to enable the network to benefit from introducing extra discriminative object information. Specifically, as shown in the top-left of \cref{fig_framework}, a larger synthetic image, created by concatenating the anchor image and another paired one, plays the role of the original image needing masking. Correspondingly, the anchor image can be treated as the erased one where the paired image is masked. Due to the introduction of extra discriminative object information, fewer object regions will be highlighted in the anchor part of the synthetic image. Therefore, we then strengthen the object localization ability of the network by transferring the more integral object knowledge from the original anchor CAM to the consequently less activated anchor part of the synthetic image. The knowledge transfer loss for the simulated inter-image erasing can be formulated as follows:
	\begin{equation}
		\begin{split}
			\mathcal{L}_{kt} =  ReLU\left (\hat{F_{a} }  -\hat{F_{s} }  \right ), 
			~~~\text{where}~ \hat{F_{a} }  =CFE\left ( F_{a}, y \right ), \hat{F_{s} }  =CFE\left ( F_{s}, y \right ).
		\end{split}
		\label{eq_local}
	\end{equation}
	Here, $F_{a}$ and $F_{s}$ denote the CAM features of the anchor and simulated branches, respectively. CFE represents class feature extraction, which extracts the feature channel related to the classes that exist in the image. The ReLU operation means that our alignment focuses on the more highlighted object region rather than the background. Note that no rotation or resize is performed for the concatenated image. We only introduce extra object information from the paired image and control other variables to guarantee the weakened activation for the anchor part. Intuitively, our knowledge transfer with simulated inter-image erasing will not cause the over-expansion problem when the object in the anchor branch is well-discovered. Our experiments demonstrate that such knowledge transfer can effectively strengthen the localization ability of the network to alleviate the under-activation, especially when multiple object instances co-exist in the image. Besides, benefiting from the bidirectional alignment, the anchor branch will also learn from the weakened activation of the simulated one and produce compact attention maps for small objects to mitigate the over-expansion problem.

	\subsection{Self-Supervised Regularization}
	Our simulated inter-image erasing aims to improve the object localization ability of the network by maintaining the high activation given extra discriminative object information. However, the adopted knowledge transfer is bidirectional, which will also weaken the anchor branch object mining when learning from the sparse activation of the simulated branch. Since simply cutting the gradient propagation of the anchor branch makes the training unstable, we propose a self-supervised regularization module for the anchor CAM feature to maintain its reliable activation in discriminative regions. Specifically, with the generated CAM feature $F_{a}$ and the corresponding CAM $A_{a}$, we first leverage two thresholds $\beta_h = 0.3$ and $\beta_l = 0.15$ to locate the confident foreground and background as follows (the subscript $a$ is omitted for simplicity):
	\begin{equation}
		\label{eq_ref_pseudo_label}
		\hat{Y}_{i,j}=\left\{
		\begin{aligned}
			& \mathtt{argmax}(F_{i,j,:}), & \text{if $\mathtt{max}(A_{i,j,:})\geq\beta_h$,} \\
			& 0,                          & \text{if $\mathtt{max}(A_{i,j,:})\leq\beta_l$,} \\
			& 255,                        & \text{otherwise,}                               \\
		\end{aligned}
		\right.
	\end{equation}
	where $255$ denote the ignored labels for uncertain regions and $\mathtt{argmax}(\cdot)$ extracts the semantic class with the maximum activation value. After refining the pseudo label $\hat{Y}$ with the pixel-adaptive refinement module \cite{ru2022learning}, we leverage it to directly supervise the learning of CAM features with the cross-entropy loss (coordinates $i,j$ are omitted):
	\begin{equation}
		\mathcal{L}_{ce}=  -  \sum_{c=0}^{C} \hat{Y}_{bg}^{c } \log  \Gamma \left ( F  \right )^{ c } + \omega _{fg}  \hat{Y}_{fg}^{c } \log  \Gamma \left ( F  \right )^{ c } .
	\end{equation}
	$\Gamma \left ( \cdot  \right )$ denotes the softmax function. We also leverage the pseudo label $\hat{Y}$ generated from the anchor branch to guide the learning of the simulated CAM feature with the same cross-entropy loss. However, we notice that the extracted labels of the foreground are quite noisy, especially for complex images with multiple categories of objects where the inter-class boundary is blurred. Here, we define images with objects of only one category as simple images and others containing two or more categories as complex ones. Therefore, we discount the influence of foreground label $\hat{Y}_{fg}$ with the weight of $\omega _{fg} = 0.0125$. Besides, as shown in the right of \cref{fig_framework}, we design an inter-class loss to implicitly encourage the activation consistency for the multiple foreground classes in complex images as follows:
	\begin{equation}\label{inter}
		\mathcal{L}_{inter}=   \frac{ \sum \left ( V^{2nd} - V^{max} \right ) \cdot M^{f}}{\left | M^{f} \right |   },
	\end{equation}
	Here, $M^{f}$ denotes the foreground mask obtained with a threshold $\beta_o = 0.2$ and $\left | \cdot  \right |$ return the value of $l_1$-norm. $V^{max}$ and $V^{2nd}$ denotes the largest and second largest activation value along the channel dimension of the CAM feature $F_{a}$, which can be obtained as follows:
	\begin{equation}
		\begin{split}
			V_{i,j}^{max} = \mathtt{max}(F_{i,j,:}), ~~~
			F^{left} =  \mathtt{remove}(F, V_{max}), ~~~
			V_{i,j}^{2nd} =  \mathtt{max}(F^{left}_{i,j,:}).
		\end{split}
		\label{eq_max}
	\end{equation}
	\cref{inter} improves the exclusivity of the class activation for each pixel, leading to more accurate boundaries between foreground objects. Our proposed self-supervised regularization module treats the simple and complex images differently, intending to maintain the reliable activation of the main branch without deteriorating the inter-class boundary caused by inaccurate pseudo labels. 
	
	\subsection{Multi-Granularity Alignment}
	Our proposed self-supervised regularization module can effectively constrain and boost the knowledge transfer in our simulated inter-image erasing module. However, the improvement of the network's object localization ability will be severely limited by the quality of the anchor CAM, which is usually under-activated. Therefore, it is natural to resort to the traditional intra-image erasing, which aims to expand the object area. Following AEFT \cite{yoon2022adversarial}, we leverage a threshold of  
	0.6 to mask the most discriminative region. To avoid introducing too much background noise, we propose a multi-granularity alignment module to gently expand the object activation, which encourages the CAM expansion of the erased image and then transfers the learned object knowledge back to the anchor branch. As shown in the bottom of \cref{fig_framework}, we first input the anchor feature $F_{a}$ and masked feature $F_{m}$ into a class feature extraction module to obtain the corresponding features of the existing target categories ($\hat{F_{a}}$ and $\hat{F_{m}}$).  Following the experimental finding of AEFT \cite{yoon2022adversarial} that rigid classification supervision \cite{zhang2018adversarial} for the masked branch tends to trigger the over-expansion problem in adversarial erasing, we resort to the soft class confidence guidance from the anchor branch to the masked one. Specifically, we directly adopt a global average pooling (GAP) operation to obtain the final class confidence for each branch and an image-level global alignment loss can be formulated as follows:
	\begin{align}
		\mathcal{L}_{global} = GAP &\left ( ReLU\left (  \hat{F_{a}}  \right )  \right ) - GAP\left ( ReLU\left ( \hat{F_{m}}  \right )  \right ),  \nonumber \\ 
		\text{where}~~~ \hat{F_{m}} &=CFE\left ( F_{m}, y \right ).
	\end{align}
	$\hat{F_{a}}$ and $CFE$ have been defined in \cref{eq_local}. Another difference with rigid classification supervision is that benefiting from the class feature extraction, our global alignment loss only focuses on the logit confidence for classes that exist in the image, enabling more effective and efficient gradient propagation. Unlike AEFT \cite{yoon2022adversarial} that utilizes the feature space of the GPP layer as embedding space for loss construction, our experiments reveal that direct alignment with the CAM feature leads to more promising performance. With class information learned from the anchor branch, we further leverage a pixel-level local activation alignment to transfer the newly discovered object information from the erased image to guide the anchor branch learning. 
	\begin{equation}
		\mathcal{L}_{local}=  ReLU\left ( \hat{F_{m}}  - \hat{F_{a}}  \right ).
		\label{eq_mlocal}
	\end{equation}
	Such pixel-level local alignment can also hinder the erased branch from activating the unwanted background area by learning from the lowly-activated regions of the anchor branch. With the gentle and constrained activation enlargement, the multi-granularity alignment module further boosts the performance of our proposed simulated inter-image erasing approach. As can be seen in \cref{fig_framework}, our proposed SIE and MGA branches are designed in a symmetric way that play the adversarial role of expanding and constraining the anchor image activation, respectively. Their synergically cooperation facilitates the convergence for mining complete and compact object regions.
	
	\subsection{Training Objective}
	The overall training loss is as follows:
	\begin{equation}
		\mathcal{L}=\mathcal{L}_{cls} + \mathcal{L}_{kt}  + \mathcal{L}_{global} + \mathcal{L}_{local} + \mathcal{L}_{ce} + \lambda_{inter}\mathcal{L}_{inter}.
		\label{eq_all}
	\end{equation}
	We empirically set $\lambda_{inter} = 0.005$ as the hyperparameter that controls the weight of inter-class loss.

	\section{Experiment}

	\subsection{Datasets and Implementation Details}
	We evaluate our approach on the PASCAL VOC 2012 \cite{everingham2010pascal} and COCO \cite{lin2014microsoft} datasets. The PASCAL VOC 2012 dataset contains 21 classes (20 object categories and the background) for semantic segmentation. It has 10,582 images for training (expanded with SBD \cite{hariharan2011semantic}), 1,449 for validation and 1,456 for testing. The COCO dataset is a more challenging benchmark with 80 semantic classes and the background. Following previous works \cite{wang2020weakly,Li2021GroupWiseSM,zhang2020causal}, we use the default train/val splits (80k images for training and 40k for validation) in the experiment. Mean intersection over union (mIoU) is adopted as the metric to evaluate the quality of our pseudo labels and segmentation results. The results for the PASCAL VOC test set are obtained from the official evaluation server.
	
	For the classification network, we follow the work of AEFT \cite{yoon2022adversarial} and employ ResNet38 \cite{wu2019wider} as the backbone. We adopt a poly
	learning rate with an initial value of $10^{-2}$ and a power of 0.9. For the second stage training of WSSS, following the recent work of BECO \cite{rong2023boundary}, we adopt DeeplabV2 \cite{chen2017deeplab} as the segmentation network, which uses ResNet101 \cite{he2016deep} as the backbone. The momentum and weight decay of the SGD optimizer are 0.9 and $10^{-4}$. The initial learning rate is set to $10^{-2}$ and is decreased using polynomial decay. The segmentation model is trained for 80 epochs and 40 epochs on VOC and COCO datasets, respectively, with a common batch size of 16. We also follow the default setting of DeeplabV2 \cite{chen2017deeplab} and conduct experiments with VGG16 backbone for a more comprehensive comparison with previous approaches \cite{fan2020learning,chen2020weakly,sun2021ecs,lee2021railroad,jiang2022l2g}. All our backbones are pre-trained on ImageNet \cite{deng2009imagenet}. 
	
	\begin{table}[t] 
		\begin{minipage}{0.48\linewidth}
			\centering
			\setlength{\tabcolsep}{1mm}
			\caption{Accuracy of pseudo-masks evaluated on PASCAL VOC 2012 training set.}
			\label{tab_pseudo}	
			
			\begin{tabular}{{lcc}}
				\hline
				Methods  & Seed &~w/~IRN \cite{ahn2019weakly} \\
				\hline
				IRN \cite{ahn2019weakly}$_{\text{\tiny{CVPR19}}}$&48.3&66.3\\
				MBMNet\cite{liu2020weakly}$_{\text{\tiny{MM20}}}$&50.2&66.8\\
				CONTA \cite{zhang2020causal}$_{\text{\tiny{NIPS20}}}$&48.8&67.9\\
				AdvCAM \cite{lee2021anti}$_{\text{\tiny{CVPR21}}}$&55.6&69.9\\
				RIB\cite{lee2021reducing}$_{\text{\tiny{NIPS21}}}$&56.6&70.6\\
				ReCAM \cite{chen2022class}$_{\text{\tiny{CVPR22}}}$&56.6&70.5\\
				ESOL\cite{li2022expansion}$_{\text{\tiny{NIPS22}}}$&53.6&68.7\\
				CLIMS\cite{xie2022clims}$_{\text{\tiny{CVPR22}}}$&56.6&70.5\\ 
				AEFT \cite{yoon2022adversarial}$_{\text{\tiny{ECCV22}}}$&56.0&71.0\\	
				ACR \cite{cheng2023out}$_{\text {\tiny{CVPR23}}}$&60.3&72.3\\    
				FPR\cite{chen2023fpr}$_{\text {\tiny{ICCV23}}}$&63.8&-\\
				\hline
				\textbf{KTSE (Ours)} &\textbf{67.0}& \textbf{73.8}\\
				\hline	
			\end{tabular}

		\end{minipage}
		\hfill
		\begin{minipage}{0.49\linewidth}
			\centering
			\setlength{\tabcolsep}{1.5mm}
			\caption{Quantitative comparisons on PASCAL VOC 2012 val and test sets with \textbf{VGG} backbone. Sup: Supervision, I: Image-level label, S: Saliency maps.}
			\label{tab_vgg}	
			
			\begin{tabular}{{l}*{3}{c}}
				\hline
				Methods &   Sup & Val & Test\\
				\hline
				DRS \cite{kim2021discriminative}$_{\text {\tiny{AAAI21}}}$&I+S&63.6&64.4\\
				GSM \cite{Li2021GroupWiseSM}$_{\text {\tiny{AAAI21}}}$&I+S&63.3&63.6\\
				NSROM \cite{yao2021non}$_{\text{\tiny{CVPR21}}}$&I+S&65.5&65.3\\
				EPS \cite{lee2021railroad}$_{\text {\tiny{CVPR21}}}$&I+S&67.0&67.3\\
				L2G \cite{jiang2022l2g}$_{\text {\tiny{CVPR22}}}$&I+S&68.5&68.9\\
				
				\hline
				
				AffinityNet\cite{ahn2018learning}$_{\text{\tiny{CVPR18}}}$&I&58.4 &60.5\\ 
				ICD \cite{fan2020learning}$_{\text{\tiny{CVPR20}}}$&I&61.2&60.9\\
				BES \cite{chen2020weakly}$_{\text{\tiny{ECCV20}}}$&I&60.1&61.1\\
				ECS \cite{sun2021ecs}$_{\text{\tiny{ICCV21}}}$&I&62.1&63.4\\	
				
				\hline
				\textbf{KTSE (Ours)} &I&\textbf{67.3} &\textbf{67.0}\\	
				
				\hline	
			\end{tabular}
			
		\end{minipage}
		\vspace{-0.15cm}
	\end{table}

	\subsection{Comparisons to the State-of-the-arts}
	
	\textbf{Accuracy of Pseudo-Masks.} We first report the quality of the segmentation seeds and the generated pseudo-masks derived from our approach. The comparison with other state-of-the-arts is presented in \cref{tab_pseudo}. As can be seen, our segmentation seed can reach the mIoU of 67.0\%, bringing a gain of 18.7\% compared to the baseline reported by IRN \cite{ahn2019weakly}. Our method can outperform the state-of-the-art method FPR \cite{chen2023fpr} by 3.2\%. With the further refinement of IRN \cite{ahn2019weakly}, the mIoU of our generated pseudo-masks can arrive at 73.8\%, surpassing the previous SOTA of AEFT \cite{yoon2022adversarial} and  ACR \cite{cheng2023out} by more than 1.5\%.

	\noindent%
	\begin{minipage}{0.49\textwidth} 
		\vspace{0pt}
		\centering
		\setlength{\tabcolsep}{1.0mm}
		\captionof{table}{Quantitative comparisons on PASCAL VOC 2012 val and test sets with \textbf{ResNet} backbone. Sup: Supervision, I: Image-level label, S: Saliency maps, O: Out-of-distribution data, L: Language.}
		\label{tab_resnet}	
		\begin{tabular}{{l}*{3}{c}}
			\hline
			Methods &   Sup &  Val & Test\\
			\hline
			DRS \cite{kim2021discriminative}$_{\text{\tiny{AAAI21}}}$&I+S&71.2&71.4\\
			NSROM \cite{yao2021non}$_{\text{\tiny{CVPR21}}}$&I+S&70.4&70.2\\
			EPS \cite{lee2021railroad}$_{\text{\tiny{CVPR21}}}$&I+S&71.0&71.8\\
			AuxSeg \cite{xu2021leveraging}$_{\text{\tiny{ICCV21}}}$&I+S&69.0&68.6\\
			PPC \cite{du2022weakly}$_{\text{\tiny{CVPR22}}}$&I+S&72.6&73.6\\
			RCA \cite{zhou2022regional}$_{\text{\tiny{CVPR22}}}$&I+S&72.2&72.8\\
			L2G \cite{jiang2022l2g}$_{\text{\tiny{CVPR22}}}$&I+S&72.1&71.7\\
			W-OoD \cite{lee2022weakly}$_{\text{\tiny{CVPR22}}}$& I+O & 69.8 & 69.9 \\
			CLIP-ES \cite{lin2023clip}$_{\text{\tiny{CVPR23}}}$&I+L&71.1&71.4\\
			LPCAM \cite{chen2023extracting}$_{\text{\tiny{CVPR23}}}$&I+S&71.8&72.1\\		  
			
			\hline   
			
			
			AdvCAM \cite{lee2021anti}$_{\text{\tiny{CVPR21}}}$&I&68.1&68.0\\
			CDA \cite{su2021context}$_{\text{\tiny{ICCV21}}}$&I&66.1&66.8\\
			CSE \cite{kweon2021unlocking}$_{\text{\tiny{ICCV21}}}$&I&68.4&68.2\\
			RIB \cite{lee2021reducing}$_{\text{\tiny{NIPS21}}}$&I&68.3&68.6\\
			AMR \cite{qin2022activation}$_{\text{\tiny{AAAI22}}}$&I&68.8&69.1\\
			MCT \cite{xu2022multi}$_{\text{\tiny{CVPR22}}}$&I&71.9&71.6\\
			AFA \cite{ru2022learning}$_{\text{\tiny{CVPR22}}}$&I& 66.0 &66.3\\
			SIPE \cite{chen2022self}$_{\text{\tiny{CVPR22}}}$& I & 68.8 & 69.7 \\
			ReCAM \cite{chen2022class}$_{\text{\tiny{CVPR22}}}$&I& 68.5 &68.4\\
			PPC \cite{du2022weakly}$_{\text{\tiny{CVPR22}}}$&I&67.7&67.4\\
			ViT-PCM \cite{rossetti2022max}$_{\text{\tiny{ECCV22}}}$&I&70.3&70.9\\
			S-BCE \cite{rossetti2022max}$_{\text{\tiny{ECCV22}}}$&I&70.0&71.3\\
			AEFT \cite{yoon2022adversarial}$_{\text{\tiny{ECCV22}}}$&I&70.9&71.7\\
			TOCO \cite{ru2023token}$_{\text{\tiny{CVPR23}}}$&I&69.8&70.5\\
			OCR \cite{cheng2023out}$_{\text{\tiny{CVPR23}}}$&I&72.7&72.0\\       	
			ACR \cite{cheng2023out}$_{\text{\tiny{CVPR23}}}$&I&71.9&71.9\\    
			BECO \cite{rong2023boundary}$_{\text{\tiny{CVPR23}}}$&I&72.1&71.8\\
			FPR \cite{chen2023fpr}$_{\text{\tiny{ICCV23}}}$&I&70.3&70.1\\
			MARS \cite{jo2023mars}$_{\text{\tiny{ICCV23}}}$&I&70.3&71.2\\
			\hline
			\textbf{KTSE (Ours)} &I&\textbf{73.0} &\textbf{72.9}\\	
			
			\hline	
		\end{tabular}  
		
	\end{minipage}
	\hfill
	\begin{minipage}{0.46\textwidth}
		\centering
		\setlength{\tabcolsep}{2.5mm}
		\captionof{table}{Quantitative comparisons on COCO val set with \textbf{VGG} backbone. Sup: Supervision, I: Image-level label, S: Saliency maps.}
		\label{tab_coco_vgg}
		\begin{tabular}{{l}*{3}{c}}
			\hline
			Methods & Sup &  Val \\
			\hline

			IAL \cite{wang2020weakly}$_{\text {\tiny{IJCV20}}}$ &I+S  &27.7\\
			GWSM \cite{Li2021GroupWiseSM}$_{\text {\tiny{AAAI21}}}$ &I+S&28.4\\
			EPS \cite{lee2021railroad}$_{\text {\tiny{CVPR21}}}$&I+S&35.7\\
			I2CRC \cite{zhou2022regional}$_{\text {\tiny{TMM22}}}$&I+S&31.2\\ 
			RCA \cite{zhou2022regional}$_{\text {\tiny{CVPR22}}}$&I+S&36.8\\
			MDBA \cite{chen2023multi}$_{\text {\tiny{TIP23}}}$&I+S&37.8\\
			\hline
			BFBP \cite{saleh2016built}$_{\text {\tiny{ECCV16}}}$&I&20.4\\ 
			SEC \cite{kolesnikov2016seed}$_{\text {\tiny{ECCV16}}}$&I&22.4\\ 
			CONTA \cite{zhang2020causal}$_{\text {\tiny{NIPS20}}}$&I &23.7\\
			\hline
			\textbf{KTSE (Ours)} &I&\textbf{37.2} \\
			\hline
		\end{tabular}

		\vspace{0.05cm}  
		\centering
		\setlength{\tabcolsep}{2.3mm}
		\captionof{table}{Quantitative comparisons on  COCO val set with \textbf{ResNet} backbone. Sup: Supervision, I: Image-level label, S: Saliency maps, L: Language.}
		\label{tab_coco_resnet}	
		\begin{tabular}{{l}*{3}{c}}
			\hline
			Methods & Sup &  Val \\
			\hline
			L2G \cite{jiang2022l2g}$_{\text{\tiny{CVPR22}}}$&I+S&44.2\\   
			CLIP-ES \cite{lin2023clip}$_{\text{\tiny{CVPR23}}}$&I+L&45.4\\
			LPCAM \cite{chen2023extracting}$_{\text{\tiny{CVPR23}}}$&I+S&42.1\\	
			\hline
					OC-CSE \cite{kweon2021unlocking}$_{\text{\tiny{ICCV21}}}$ &I &36.4\\
					CDA \cite{su2021context}$_{\text{\tiny{ICCV21}}}$&I&33.2\\
					RIB \cite{lee2021reducing}$_{\text{\tiny{NIPS21}}}$&I&43.8\\
					MCT \cite{xu2022multi}$_{\text{\tiny{CVPR22}}}$&I&42.0\\
					SIPE \cite{chen2022self}$_{\text{\tiny{CVPR22}}}$& I & 40.6 \\
					TOCO \cite{ru2023token}$_{\text{\tiny{CVPR23}}}$&I&41.3\\
					OCR \cite{cheng2023out}$_{\text{\tiny{CVPR23}}}$&I&42.5\\    
					ACR \cite{cheng2023out}$_{\text{\tiny{CVPR23}}}$&I&45.3\\  
					BECO \cite{rong2023boundary}$_{\text{\tiny{CVPR23}}}$&I&45.1\\
					FPR \cite{chen2023fpr}$_{\text{\tiny{ICCV23}}}$&I&43.9\\
					USAGE \cite{peng2023usage}$_{\text{\tiny{ICCV23}}}$&I&44.3\\  
					\hline
					\textbf{KTSE (Ours)} &I&\textbf{45.9} \\
					\hline
				\end{tabular}  
				
			\end{minipage}	
			\vspace{0.5cm}

			\noindent\textbf{Accuracy of Segmentation Maps on PASCAL VOC 2012.} Our segmentation results on the PASCAL VOC 2012 dataset with the backbone of VGG and ResNet are demonstrated in \cref{tab_vgg} and \cref{tab_resnet}, respectively. As can be seen, with the VGG backbone, our approach achieves the performance of 67.3\% on the validation set and 67.0\% on the test set,  better than other state-of-the-art methods with only image-level labels. Our segmentation results are also competitive to many approaches that rely on saliency maps, \eg, NSROM \cite{yao2021non} and EPS \cite{lee2021railroad}. With the ResNet backbone, we can improve the results to 73.0\% and 72.9\% on the validation and test sets, respectively. As can be seen in \cref{tab_resnet}, our proposed approach can achieve better performance compared with recent SOTA methods. For example, our approach outperforms OCR \cite{cheng2023out} and ACR \cite{cheng2023out} by about 1\% on the test set, demonstrating the superiority of our knowledge transfer with simulated inter-image erasing.

			\noindent\textbf{Accuracy of Segmentation Maps on COCO.} For the more challenging COCO dataset, we provide performance comparisons with state-of-the-art WSSS methods using the backbone of VGG and ResNet in \cref{tab_coco_vgg} and \cref{tab_coco_resnet}, respectively. As shown in \cref{tab_coco_vgg}, our proposed KTSE with VGG backbone can achieve the performance of 37.2\% mIoU, much better than previous methods supervised with only image-level labels, \eg, 13.5\% mIoU higher than CONTA \cite{zhang2020causal}. Besides, our results are also competitive compared to previous SOTA methods with additional saliency guidance like RCA \cite{zhou2022regional} and MDBA \cite{chen2023multi}. With the ResNet backbone, our proposed KTSE reaches the best result of 45.9\% mIoU compared to previous SOTA WSSS methods. Specifically, our approach outperforms ACR \cite{cheng2023out} and BECO \cite{rong2023boundary} by 0.6\% and 0.8\% mIoU, respectively.

			\begin{table}[t] 
				\begin{minipage}{0.5\linewidth}
					\centering
					\setlength{\tabcolsep}{2.5mm}
					\caption{Element-wise component analysis. SIE: Simulated Inter-Image Erasing, SSR: Self-Supervised Regularization, MGA: Multi-Granularity Alignment. }
					\label{tab_element}
					
					\begin{tabular}{*{5}{c}}
						\hline
						Base&SIE&SSR&MGA&mIoU\\
						\hline
						\checkmark&&&&54.2\\
						\checkmark&\checkmark&&&57.1\\
						\checkmark&&\checkmark&&57.4\\
						\checkmark&&&\checkmark&60.2\\
						\checkmark&\checkmark&\checkmark&&63.8\\
						\checkmark&\checkmark&\checkmark&\checkmark&\textbf{67.0}\\	
						
						\hline
					\end{tabular}
					
				\end{minipage}
				\hfill
				\begin{minipage}{0.45\linewidth}
					\centering
					\setlength{\tabcolsep}{3mm}
					\renewcommand\arraystretch{1.15}
					\caption{Comparison of Our Multi-Granularity Alignment with Previous Erasing-based Approaches. GA: Global Alignment; LA: Local Alignment.}
					\label{tab_ab}
					
					\begin{tabular}{*{1}{l}{c}}
						
						\hline
						Method&mIoU\\
						\hline
						Base&54.2\\
						Rigid Classification \cite{zhang2018adversarial} &55.6\\
						Soft GPP Feature \cite{yoon2022adversarial} &54.9\\
						\textbf{Our GA} &\textbf{58.7}\\
						\hline
						\textbf{Our GA \& LA} &\textbf{60.2}\\
						\hline
					\end{tabular}
					
				\end{minipage}
				
			\end{table}

			\subsection{Ablation Studies}
			\textbf{Element-Wise Component Analysis.}
			In this part, we demonstrate the contribution of each component proposed in our approach to improving the quality of pseudo-masks. As mentioned in \cref{cam_g}, we incorporate the GPP \cite{yoon2022adversarial} into the classification network as our strong baseline. As shown in \cref{tab_element}, with our simulated inter-image erasing (SIE) module, we can improve the accuracy of segmentation seeds from the baseline of 54.2\% to 57.1\%, which demonstrates the benefit of strengthening the localization ability of the network through weakening the activation with extra object information and then learning from the anchor branch. By maintaining reliable activation in discriminative regions, our self-supervised regularization (SSR) module can achieve the mIoU of 57.4\%. Moreover, the combination of our proposed SIE and SSR can significantly boost the performance to 63.8\% mIoU. With our proposed multi-granularity alignment module to gently expand the object activation, the mIoU of pseudo masks finally arrives at 67.0\%, highlighting the importance of improving the quality of anchor CAM to boost the performance of our knowledge transfer with simulated inter-image erasing.

\begin{figure*}[t]
				\centering
				\includegraphics[width=1.0\linewidth]{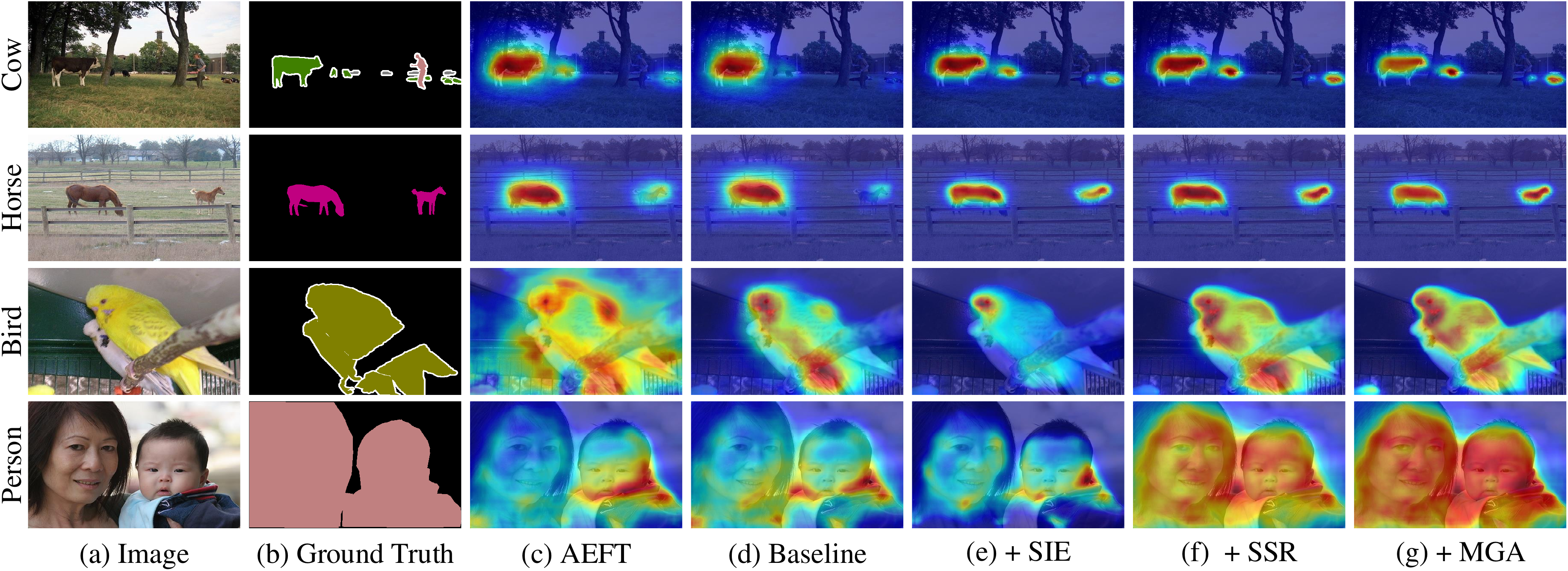}
				\caption{Example localization maps on the PASCAL VOC 2012 training set. For each (a) image, we show (b) ground truth, localization maps produced by (c) previous work of AEFT \cite{yoon2022adversarial}, (d) our baseline, (e) baseline + SIE, (f) baseline + SIE + SSR, and (g) baseline + SIE + SSR + MGA. Best viewed in color.}
				\label{fig_cam}
			\end{figure*}

			Some example localization maps on the PASCAL VOC 2012 training set can be viewed in \cref{fig_cam}. As seen from the first two rows, after strengthening the localization ability of the network with our simulated inter-image erasing (SIE) module, we can successfully expand the object activation and discover the less discriminative object in the image (\eg, the smaller animals on the right). We can notice that our SIE can also help alleviate the over-expansion problem and generate more compact activation, \eg, the cow in the first row. However, as demonstrated in the last two rows, the bidirectional alignment in SIE will also lead to more severe under-activation when the anchor branch learns from the sparse activation of the simulated one, especially for larger objects. Fortunately, our proposed self-supervised regularization (SSR) module can help maintain reliable activation in discriminative regions to guarantee effective knowledge transfer. Moreover, as seen in the last row, our multi-granularity alignment (MGA) module can further help discover the integral target region by gently expanding the object activation. Comparing the results of ours with that of AEFT \cite{yoon2022adversarial},  \cref{fig_cam} (g) vs  \cref{fig_cam} (c), we can see that our proposed approach can significantly alleviate the under-activation and over-expansion of previous approaches due to insufficient or excessive erasure.

			\noindent \textbf{MGA \vs Previous Erasing-based Approaches.}
			\label{sec:dis}
			For the multi-granularity alignment, we resort to the soft class confidence knowledge from the anchor branch to guide the gentle activation expansion of the masked image. In \cref{tab_ab}, we compare performance improvement from different guidance settings. As can be seen, our global alignment of CAM feature can improve the baseline from 54.2\% to 58.7\%, outperforming the rigid classification guidance in ACoL \cite{zhang2018adversarial} and GPP feature alignment adopted in AEFT \cite{yoon2022adversarial} by 3.1\% and 3.8\%, respectively. It highlights the importance of direct global alignment in CAM features rather than embedding space with extra projection layers. With our pixel-level local alignment, our multi-granularity alignment module can finally improve the performance to 60.2\%, demonstrating the benefit of our gentle alignment strategy compared with previous erasing-based approaches.

			\noindent\textbf{Phenomenon of SIE.} 
			Based on the observation of classic adversarial erasing methods that removing the most discriminative region will lead to the extra activation of other object parts, the reversal thinking comes naturally--what if we introduce extra discriminative object information? The answer is the original highly activated area will become less discriminative and be lowly activated. We then enhance this weakened activation by learning from the original CAM. When the network learns to increase the attention for the less-discriminative region in the concatenated image, it will also learn to activate less-discriminative object region in the original image to locate more objects. 
			
			\noindent\textbf{SIE \vs Data Augmentation.} Though data augmentations like CutMix will also change the anchor image and lead to activation perturbation, they do not guarantee the weakened activation (which might cause over-expansion) for the anchor part of the concatenated image like ours. In contrast, we only introduce extra object information from the paired image and control other variables. Most importantly, our novelty lies in constructing the simulated inter-image erasing scenario (reversal thinking on existing masking methods) rather than simply augmenting the data input. It enhances the network's localization ability by improving the consequent less activated attention map through learning from the object knowledge of the anchor branch. Accompanied by our proposed MGA, they are designed in a symmetric way that play the adversarial role of expanding and constraining the anchor image activation respectively for mining complete and compact object regions.  Note that our approach significantly outperforms the CutMix-based augmentation method of CDA \cite{su2021context} with 73.0 \vs 66.1 on VOC (\cref{tab_resnet}) and 45.9 \vs 33.2 on COCO (\cref{tab_coco_resnet}).

			\section{Conclusion}
			This paper proposed a Knowledge Transfer with Simulated Inter-Image Erasing (KTSE) approach for weakly supervised semantic segmentation. Specifically, in contrast to existing adversarial erasing-based methods that remove the discriminative part for mining less-discriminative areas, we proposed to simulate an inter-image erasing scenario where extra object information was added through concatenating a paired image. We then strengthened the object localization ability of the network by enhancing the consequent less activated localization map. In addition, we proposed a self-supervised regularization module to maintain reliable activation in the discriminative regions and improve the inter-class object boundary recognition for complex images. Moreover, we proposed a multi-granularity alignment module to gently enlarge the object activation via intra-image erasing for boosting the object knowledge transfer. Extensive experiments and ablation studies were conducted on PASCAL VOC 2012 and COCO datasets to demonstrate the superiority of our proposed approach.

			\bibliographystyle{splncs04}
			\bibliography{egbib}

\begin{thebibliography}{10}
\providecommand{\url}[1]{\texttt{#1}}
\providecommand{\urlprefix}{URL }
\providecommand{\doi}[1]{https://doi.org/#1}

\bibitem{ahn2019weakly}
Ahn, J., Cho, S., Kwak, S.: Weakly supervised learning of instance segmentation
  with inter-pixel relations. In: CVPR. pp. 2209--2218 (2019)

\bibitem{ahn2018learning}
Ahn, J., Kwak, S.: Learning pixel-level semantic affinity with image-level
  supervision for weakly supervised semantic segmentation. In: CVPR. pp.
  4981--4990 (2018)

\bibitem{bearman2016s}
Bearman, A., Russakovsky, O., Ferrari, V., Fei-Fei, L.: What's the point:
  Semantic segmentation with point supervision. In: ECCV. pp. 549--565 (2016)

\bibitem{cai2024poly}
Cai, X., Lai, Q., Wang, Y., Wang, W., Sun, Z., Yao, Y.: Poly kernel inception
  network for remote sensing detection. In: CVPR. pp. 27706--27716 (2024)

\bibitem{chen2017deeplab}
Chen, L.C., Papandreou, G., Kokkinos, I., Murphy, K., Yuille, A.: Deeplab:
  Semantic image segmentation with deep convolutional nets, atrous convolution,
  and fully connected crfs. In: IEEE TPAMI. vol.~40, pp. 834--848 (2017)

\bibitem{chen2023fpr}
Chen, L., Lei, C., Li, R., Li, S., Zhang, Z., Zhang, L.: Fpr: False positive
  rectification for weakly supervised semantic segmentation. In: ICCV. pp.
  1108--1118 (2023)

\bibitem{chen2020weakly}
Chen, L., Wu, W., Fu, C., Han, X., Zhang, Y.T.: Weakly supervised semantic
  segmentation with boundary exploration. In: ECCV (2020)

\bibitem{chen2022self}
Chen, Q., Yang, L., Lai, J.H., Xie, X.: Self-supervised image-specific
  prototype exploration for weakly supervised semantic segmentation. In: CVPR.
  pp. 4288--4298 (2022)

\bibitem{chen2021semantically}
Chen, T., Xie, G.S., Yao, Y., Wang, Q., Shen, F., Tang, Z., Zhang, J.:
  Semantically meaningful class prototype learning for one-shot image
  segmentation. IEEE TMM  \textbf{24},  968--980 (2021)

\bibitem{chen2024spatial}
Chen, T., Yao, Y., Huang, X., Li, Z., Nie, L., Tang, J.: Spatial structure
  constraints for weakly supervised semantic segmentation. IEEE TIP
  \textbf{33},  1136--1148 (2024)

\bibitem{chen2023multi}
Chen, T., Yao, Y., Tang, J.: Multi-granularity denoising and bidirectional
  alignment for weakly supervised semantic segmentation. IEEE TIP  \textbf{32},
   2960--2971 (2023)

\bibitem{chen2022saliency}
Chen, T., Yao, Y., Zhang, L., Wang, Q., Xie, G.S., Shen, F.: Saliency guided
  inter-and intra-class relation constraints for weakly supervised semantic
  segmentation. IEEE TMM  \textbf{25},  1727--1737 (2022)

\bibitem{chen2023extracting}
Chen, Z., Sun, Q.: Extracting class activation maps from non-discriminative
  features as well. In: CVPR. pp. 3135--3144 (2023)

\bibitem{chen2022class}
Chen, Z., Wang, T., Wu, X., Hua, X.S., Zhang, H., Sun, Q.: Class re-activation
  maps for weakly-supervised semantic segmentation. In: CVPR. pp. 969--978
  (2022)

\bibitem{cheng2023out}
Cheng, Z., Qiao, P., Li, K., Li, S., Wei, P., Ji, X., Yuan, L., Liu, C., Chen,
  J.: Out-of-candidate rectification for weakly supervised semantic
  segmentation. In: CVPR. pp. 23673--23684 (2023)

\bibitem{dai2015boxsup}
Dai, J., He, K., Sun, J.: Boxsup: Exploiting bounding boxes to supervise
  convolutional networks for semantic segmentation. In: ICCV. pp. 1635--1643
  (2015)

\bibitem{deng2009imagenet}
Deng, J., Dong, W., Socher, R., Li, L.J., Li, K., Fei-Fei, L.: Imagenet: A
  large-scale hierarchical image database. In: CVPR. pp. 248--255 (2009)

\bibitem{du2022weakly}
Du, Y., Fu, Z., Liu, Q., Wang, Y.: Weakly supervised semantic segmentation by
  pixel-to-prototype contrast. In: CVPR. pp. 4320--4329 (2022)

\bibitem{everingham2010pascal}
Everingham, M., Gool, L.V., Williams, C.K., Winn, J., Zisserman, A.: The pascal
  visual object classes (voc) challenge. In: IJCV. vol.~88, pp. 303--338 (2010)

\bibitem{fan2020learning}
Fan, J., Zhang, Z., Song, C., Tan, T.: Learning integral objects with
  intra-class discriminator for weakly-supervised semantic segmentation. In:
  CVPR. pp. 4283--4292 (2020)

\bibitem{fan2020cian}
Fan, J., Zhang, Z., Tan, T., Song, C., Xiao, J.: Cian: Cross-image affinity net
  for weakly supervised semantic segmentation. In: AAAI. vol.~34, pp.
  10762--10769 (2020)

\bibitem{hariharan2011semantic}
Hariharan, B., Arbel{\'a}ez, P., Bourdev, L.D., Maji, S., Malik, J.: Semantic
  contours from inverse detectors. In: ICCV. pp. 991--998 (2011)

\bibitem{he2016deep}
He, K., Zhang, X., Ren, S., Sun, J.: Deep residual learning for image
  recognition. In: CVPR. pp. 770--778 (2016)

\bibitem{jiang2022l2g}
Jiang, P.T., Yang, Y., Hou, Q., Wei, Y.: L2g: A simple local-to-global
  knowledge transfer framework for weakly supervised semantic segmentation. In:
  CVPR. pp. 16886--16896 (2022)

\bibitem{jing2019coarse}
Jing, L., Chen, Y., Tian, Y.: Coarse-to-fine semantic segmentation from
  image-level labels. IEEE TIP  \textbf{29},  225--236 (2019)

\bibitem{jo2023mars}
Jo, S., Yu, I.J., Kim, K.: Mars: Model-agnostic biased object removal without
  additional supervision for weakly-supervised semantic segmentation. arXiv
  preprint arXiv:2304.09913  (2023)

\bibitem{khoreva2017simple}
Khoreva, A., Benenson, R., Hosang, J., Hein, M., Schiele, B.: Simple does it:
  Weakly supervised instance and semantic segmentation. In: CVPR. pp. 876--885
  (2017)

\bibitem{kim2021discriminative}
Kim, B., Han, S., Kim, J.: Discriminative region suppression for
  weakly-supervised semantic segmentation. In: AAAI. vol.~35, pp. 1754--1761
  (2021)

\bibitem{kolesnikov2016seed}
Kolesnikov, A., Lampert, C.: Seed, expand and constrain: Three principles for
  weakly-supervised image segmentation. In: ECCV. pp. 695--711 (2016)

\bibitem{kweon2021unlocking}
Kweon, H., Yoon, S.H., Kim, H., Park, D., Yoon, K.J.: Unlocking the potential
  of ordinary classifier: Class-specific adversarial erasing framework for
  weakly supervised semantic segmentation. In: ICCV. pp. 6994--7003 (2021)

\bibitem{lee2021reducing}
Lee, J., Choi, J., Mok, J., Yoon, S.: Reducing information bottleneck for
  weakly supervised semantic segmentation. In: NeurIPS. vol.~34, pp.
  27408--27421 (2021)

\bibitem{lee2021anti}
Lee, J., Kim, E., Yoon, S.: Anti-adversarially manipulated attributions for
  weakly and semi-supervised semantic segmentation. In: CVPR. pp. 4071--4080
  (2021)

\bibitem{lee2022weakly}
Lee, J., Oh, S.J., Yun, S., Choe, J., Kim, E., Yoon, S.: Weakly supervised
  semantic segmentation using out-of-distribution data. In: CVPR. pp.
  16897--16906 (2022)

\bibitem{lee2021railroad}
Lee, S., Lee, M., Lee, J., Shim, H.: Railroad is not a train: Saliency as
  pseudo-pixel supervision for weakly supervised semantic segmentation. In:
  CVPR. pp. 5495--5505 (2021)

\bibitem{li2022expansion}
Li, J., Jie, Z., Wang, X., Wei, X., Ma, L.: Expansion and shrinkage of
  localization for weakly-supervised semantic segmentation. vol.~35, pp.
  16037--16051 (2022)

\bibitem{Li2021GroupWiseSM}
Li, X., Zhou, T., Li, J., Zhou, Y., Zhang, Z.: Group-wise semantic mining for
  weakly supervised semantic segmentation. In: AAAI. pp. 1984--1992 (2021)

\bibitem{lin2016scribblesup}
Lin, D., Dai, J., Jia, J., He, K., Sun, J.: Scribblesup: Scribble-supervised
  convolutional networks for semantic segmentation. In: CVPR. pp. 3159--3167
  (2016)

\bibitem{lin2014microsoft}
Lin, T.Y., Maire, M., Belongie, S., Hays, J., Perona, P., Ramanan, D.,
  Doll{\'a}r, P., Zitnick, L.: Microsoft coco: Common objects in context. In:
  ECCV. pp. 740--755 (2014)

\bibitem{lin2023clip}
Lin, Y., Chen, M., Wang, W., Wu, B., Li, K., Lin, B., Liu, H., He, X.: Clip is
  also an efficient segmenter: A text-driven approach for weakly supervised
  semantic segmentation. In: CVPR. pp. 15305--15314 (2023)

\bibitem{liu2020weakly}
Liu, W., Zhang, C., Lin, G., Hung, T.Y., Miao, C.: Weakly supervised
  segmentation with maximum bipartite graph matching. In: ACM MM. pp.
  2085--2094 (2020)

\bibitem{long2015fully}
Long, J., Shelhamer, E., Darrell, T.: Fully convolutional networks for semantic
  segmentation. In: CVPR. pp. 3431--3440 (2015)

\bibitem{pei2024videomac}
Pei, G., Chen, T., Jiang, X., Liu, H., Sun, Z., Yao, Y.: Videomac: Video masked
  autoencoders meet convnets. In: CVPR. pp. 22733--22743 (2024)

\bibitem{peng2023usage}
Peng, Z., Wang, G., Xie, L., Jiang, D., Shen, W., Tian, Q.: Usage: A unified
  seed area generation paradigm for weakly supervised semantic segmentation.
  arXiv preprint arXiv:2303.07806  (2023)

\bibitem{qin2022activation}
Qin, J., Wu, J., Xiao, X., Li, L., Wang, X.: Activation modulation and
  recalibration scheme for weakly supervised semantic segmentation. In: AAAI.
  vol.~36, pp. 2117--2125 (2022)

\bibitem{rong2023boundary}
Rong, S., Tu, B., Wang, Z., Li, J.: Boundary-enhanced co-training for weakly
  supervised semantic segmentation. In: CVPR. pp. 19574--19584 (2023)

\bibitem{rossetti2022max}
Rossetti, S., Zappia, D., Sanzari, M., Schaerf, M., Pirri, F.: Max pooling with
  vision transformers reconciles class and shape in weakly supervised semantic
  segmentation. In: ECCV. pp. 446--463. Springer (2022)

\bibitem{ru2022learning}
Ru, L., Zhan, Y., Yu, B., Du, B.: Learning affinity from attention: End-to-end
  weakly-supervised semantic segmentation with transformers. In: CVPR. pp.
  16846--16855 (2022)

\bibitem{ru2023token}
Ru, L., Zheng, H., Zhan, Y., Du, B.: Token contrast for weakly-supervised
  semantic segmentation. In: CVPR. pp. 3093--3102 (2023)

\bibitem{saleh2016built}
Saleh, F., Aliakbarian, M.S., Salzmann, M., Petersson, L., Gould, S., Alvarez,
  J.M.: Built-in foreground/background prior for weakly-supervised semantic
  segmentation. In: ECCV. pp. 413--432 (2016)

\bibitem{sheng2024adaptive}
Sheng, M., Sun, Z., Cai, Z., Chen, T., Zhou, Y., Yao, Y.: Adaptive integration
  of partial label learning and negative learning for enhanced noisy label
  learning. In: AAAI. vol.~38, pp. 4820--4828 (2024)

\bibitem{shimoda2019self}
Shimoda, W., Yanai, K.: Self-supervised difference detection for
  weakly-supervised semantic segmentation. In: ICCV. pp. 5208--5217 (2019)

\bibitem{singh2018hide}
Singh, K.K., Yu, H., Sarmasi, A., Pradeep, G., Lee, Y.J.: Hide-and-seek: A data
  augmentation technique for weakly-supervised localization and beyond. arXiv
  preprint arXiv:1811.02545  (2018)

\bibitem{song2019box}
Song, C., Huang, Y., Ouyang, W., Wang, L.: Box-driven class-wise region masking
  and filling rate guided loss for weakly supervised semantic segmentation. In:
  CVPR. pp. 3136--3145 (2019)

\bibitem{su2021context}
Su, Y., Sun, R., Lin, G., Wu, Q.: Context decoupling augmentation for weakly
  supervised semantic segmentation. In: ICCV (2021)

\bibitem{sun2021ecs}
Sun, K., Shi, H., Zhang, Z., Huang, Y.: Ecs-net: Improving weakly supervised
  semantic segmentation by using connections between class activation maps. In:
  ICCV. pp. 7283--7292 (2021)

\bibitem{vernaza2017learning}
Vernaza, P., Chandraker, M.: Learning random-walk label propagation for
  weakly-supervised semantic segmentation. In: CVPR. pp. 7158--7166 (2017)

\bibitem{wang2022looking}
Wang, W., Sun, G., Van~Gool, L.: Looking beyond single images for weakly
  supervised semantic segmentation learning. IEEE TPAMI  (2022)

\bibitem{wang2020weakly}
Wang, X., Liu, S., Ma, H., Yang, M.H.: Weakly-supervised semantic segmentation
  by iterative affinity learning. vol.~128, pp. 1736--1749. Springer (2020)

\bibitem{wang2020self}
Wang, Y., Zhang, J., Kan, M., Shan, S., Chen, X.: Self-supervised equivariant
  attention mechanism for weakly supervised semantic segmentation. In: CVPR.
  pp. 12275--12284 (2020)

\bibitem{wei2017object}
Wei, Y., Feng, J., Liang, X., Cheng, M.M., Zhao, Y., Yan, S.: Object region
  mining with adversarial erasing: A simple classification to semantic
  segmentation approach. In: CVPR. pp. 1568--1576 (2017)

\bibitem{wei2018revisiting}
Wei, Y., Xiao, H., Shi, H., Jie, Z., Feng, J., Huang, T.: Revisiting dilated
  convolution: A simple approach for weakly-and semi-supervised semantic
  segmentation. In: CVPR. pp. 7268--7277 (2018)

\bibitem{wu2019wider}
Wu, Z., Shen, C., Van Den~Hengel, A.: Wider or deeper: Revisiting the resnet
  model for visual recognition. PR  \textbf{90},  119--133 (2019)

\bibitem{xie2022clims}
Xie, J., Hou, X., Ye, K., Shen, L.: Clims: Cross language image matching for
  weakly supervised semantic segmentation. In: CVPR. pp. 4483--4492 (2022)

\bibitem{xu2021leveraging}
Xu, L., Ouyang, W., Bennamoun, M., Boussaid, F., Sohel, F., Xu, D.: Leveraging
  auxiliary tasks with affinity learning for weakly supervised semantic
  segmentation. In: ICCV. pp. 6984--6993 (2021)

\bibitem{xu2022multi}
Xu, L., Ouyang, W., Bennamoun, M., Boussaid, F., Xu, D.: Multi-class token
  transformer for weakly supervised semantic segmentation. In: CVPR. pp.
  4310--4319 (2022)

\bibitem{yao2021non}
Yao, Y., Chen, T., Xie, G.S., Zhang, C., Shen, F., Wu, Q., Tang, Z., Zhang, J.:
  Non-salient region object mining for weakly supervised semantic segmentation.
  In: CVPR. pp. 2623--2632 (2021)

\bibitem{yoon2022adversarial}
Yoon, S.H., Kweon, H., Cho, J., Kim, S., Yoon, K.J.: Adversarial erasing
  framework via triplet with gated pyramid pooling layer for weakly supervised
  semantic segmentation. In: ECCV. pp. 326--344. Springer (2022)

\bibitem{zhang2020causal}
Zhang, D., Zhang, H., Tang, J., Hua, X.S., Sun, Q.: Causal intervention for
  weakly-supervised semantic segmentation. In: NeurIPS. vol.~33 (2020)

\bibitem{zhang2018adversarial}
Zhang, X., Wei, Y., Feng, J., Yang, Y., Huang, T.S.: Adversarial complementary
  learning for weakly supervised object localization. In: CVPR. pp. 1325--1334
  (2018)

\bibitem{zhong2020random}
Zhong, Z., Zheng, L., Kang, G., Li, S., Yang, Y.: Random erasing data
  augmentation. In: AAAI. vol.~34, pp. 13001--13008 (2020)

\bibitem{zhou2016learning}
Zhou, B., Khosla, A., Lapedriza, {\`A}., Oliva, A., Torralba, A.: Learning deep
  features for discriminative localization. In: CVPR. pp. 2921--2929 (2016)

\bibitem{zhou2021group}
Zhou, T., Li, L., Li, X., Feng, C.M., Li, J., Shao, L.: Group-wise learning for
  weakly supervised semantic segmentation. IEEE TIP  \textbf{31},  799--811
  (2021)

\bibitem{zhou2022regional}
Zhou, T., Zhang, M., Zhao, F., Li, J.: Regional semantic contrast and
  aggregation for weakly supervised semantic segmentation. In: CVPR. pp.
  4299--4309 (2022)

\end{thebibliography}
		\end{document}


In this supplementary material, we provide the algorithm and additional evaluations of our proposed SED.
We further investigate the effect of the detailed composition of the SCS and SCR in our SED.
Additional experiments are conducted and reported according to the following themes:
\begin{itemize}
    \item \textbf{The Proposed SED Algorithm.} Considering to further deepen the understanding of our proposed SED, we provide the SED algorithm in Algorithm \ref{alg}. 
Details of our SED are shown in Fig.{\color{red}2} and Algorithm \ref{alg}.
    \item \textbf{Local and Global Threshold in SCS.} Considering class balance during sample selection, our SCS adaptively adjusts the threshold in an epoch-wise and class-wise manner to enable effective clean sample identification. Table~\ref{table:1} shows the results of using SCS with and without local and global thresholds.
    \item \textbf{Truncated Normal Distribution in SCR.} Considering the confidence of the corrected label of noisy samples, we propose a dynamic truncated normal distribution as a sample re-weighting function to mitigate the biased label correction. Table~\ref{table:1} also shows results of re-weighting and non-reweighting experimental setups.
    \item \textbf{Effect of Hyper-parameters in EMA.} To dynamically reflect the learning performance changes of the model, we use the exponential moving average (EMA) to dynamically update self-adaptive thresholds in SCS, truncated normal distribution parameters in SCR, and the mean-teacher model ($\theta^*$). The results of the ablation experiments are presented in Table~\ref{table:2} and Table~\ref{table:3}.
    \item \textbf{Extend Experiment Results.} To further verify the robustness of our SED as the training processes, Fig.~\ref{figure1} and Fig.~\ref{figure2} record the test accuracy of the training process under different noise conditions on CIFAR100N and CIFAR80N. Fig.~\ref{figure3}, Fig.~\ref{figure4}, and Fig.~\ref{figure5} present some sample selection results of our SED on real-world datasets.
\end{itemize}
\section{The Proposed SED Algorithm}
In summary, our proposed SED follows the paradigm that integrates sample selection, label correction, and sample reweighting for addressing noisy labels. Details of our SED
are shown in  Algorithm \ref{alg}.
We first warm up the network for $E_w$ epochs without any robust methods.
Then, we obtain the global and local thresholds and divide the clean subset $D_{c}$ and the noisy set $D_{noisy}$ in a self-adaptive and class-balanced manner by Eqs. (1-4).
For samples in the noisy subset $D_{c}$, we calculate the classification loss $\mathcal{L}_{D_{c}}$ with the given label.
For samples in the noisy subset $D_{noisy}$, we discard their given labels and perform label correction based on a mean-teacher model.
Besides, we adaptively assign different weights to samples according to their correction confidence by Eqs. (5) and calculate the classification loss $\mathcal{L}_{D_{c}}$.
Finally, we incorporate an additional weighted classification loss on clean samples $\mathcal{L}_{D_{reg}}$ \wrt corrected labels (similar to $\mathcal{L}_{D_{n}}$) to further enhance the robustness of the model.

\begin{algorithm}[t]
\caption{Our proposed algorithm}
\begin{flushleft}
\textbf{Input:} The training set $D$, the robust and non-robust networks $\theta_A$ and $\theta_B$, warm-up epochs $E_{w}$, total epochs $E_{total}$, batch size $bs$.
\end{flushleft}
\begin{algorithmic}[1] 
    \FOR {$epoch=1,2,\ldots,E_{total}$}
    \IF {$epoch \le E_{w}$}
    \FOR {$iteration=1,2,\ldots$}
    \STATE Fetch $B=\{(x_i, y_i)\}^{bs}$ from $D$;
    \STATE Calculate $\mathcal{L}_{\theta_A} = - \sum_{(x,y) \in B}y\ \mathcal{\log}\ F(x,\theta_A)$;
    \STATE Calculate $\mathcal{L}_{\theta_B} = - \sum_{(x,y) \in B}y\ \mathcal{\log}\ F(x,\theta_A)$;
    \STATE Update $\theta_A$ and $\theta_B$ by optimizing $\mathcal{L}_{\theta_A}$ and $\mathcal{L}_{\theta_B}$.
    \ENDFOR
    \ENDIF
    
    \IF {$E_{w} < epoch \le E_{total}$}
    \FOR {$iteration=1,2,\ldots$}
    \STATE Select ``clean'' samples using Eq. (6);
    \STATE Mining more ``clean'' samples using Eq. (8);
    \STATE Calculate the reliability of $D_c$ using Eq. (9);
    \STATE Calculate $\mathcal{L}_{\theta_A}$ using Eq. (11);
    \STATE Calculate $\mathcal{L}_{\theta_B} = - \sum_{(x,y) \in B}y\ \mathcal{\log}\ F(x,\theta_A)$;
    \STATE Update $\theta_A$ and $\theta_B$ by optimizing $\mathcal{L}_{\theta_A}$ and $\mathcal{L}_{\theta_B}$.
    \ENDFOR
    \ENDIF
    \ENDFOR
\end{algorithmic}
\begin{flushleft}
\textbf{Output:} The updated robust network $\theta_A$.
\end{flushleft}
\label{alg}
\end{algorithm}

\section{Futher Analysis}
In this section, we investigate the impact of the detailed composition of the SCS and SCR in our SED.
As outlined, our proposed SED selects and re-weights samples based on class-specific thresholds that are calculated in a data-driven manner. 
This helps to promote self-adaptivity and balance in both sample selection and re-weighting.
Specifically, we mainly explore the effect of local thresholds in our SCS and the dynamic truncated normal distribution in our SCR during the training process.
Besides, we employ the exponential moving average (EMA) to further refine these thresholds to alleviate unstable training.
Thus, we further explore the influence of the hyper-parameter (\ie, $m$ and $\alpha$) of the exponential moving average in SCS, SCR, and the mean-teacher model $\theta^*$.

\begin{table}[t]
\vspace{+0.38cm}
\renewcommand\tabcolsep{5pt}
\centering
\begin{tabular}{l|c}
\toprule
\textbf{Model} & \textbf{Test Accuracy}\\
\midrule
Standard & 34.10\\
\midrule
Standard+SCS w/o local threshold & 53.36\\
Standard+SCS w/o global threshold & 55.64\\
Standard+SCS w/o EMA & 54.72\\
Standard+SCS & 58.21\\
Standard+SCR & 54.42\\
Standard+CR & 55.51\\
Standard+SCS+SCR w/o re-weighting & 59.75\\
Standard+SCS+SCR w/o EMA & 60.08\\
Standard+SCS+SCR & 60.43\\
Standard+SCS+CR & 59.28\\
Standard+SCR+CR & 55.98\\
Standard+SCS+SCR+CR & 62.65\\
\bottomrule
\end{tabular}
\caption{Effect of each component in test accuracy (\%) on CIFAR100N (Sym-50\%).}
\label{table:1}
\end{table}

\subsection{ Local and Global Thresholds in SCS. }
Our proposed SCS adaptively adjusts the threshold in an epoch-wise and class-wise manner to enable effective clean sample identification.
As shown in Table~\ref{table:1}, we provide the result of using SCS without local and global thresholds. 
The performance drops by 4.85\% and 2.57\% accordingly.
This proves that our local threshold design is crucial for improving the robustness of the model.
By considering class balance during sample selection, our SCS can significantly alleviate the learning bias of the model.

Furthermore, our SCS employs the exponential moving average (EMA) \cite{josrc} to further refine the global and local threshold.
This helps to alleviate the issue of unstable training caused by large perturbation of the averaged predicted probability.
As indicated in Table~\ref{table:1}, it is noticeable that the utilization of EMA has resulted in a significant increase (\ie, 3.49\%) in model performance. 
This clearly demonstrates the advantageous impact of EMA on our SCS.


\begin{figure*}[t]
	\centering
	\includegraphics[width=0.8\linewidth]{supplementary/supplementary_2.pdf}
	\caption{The overall precision of sample selection (\%) \vs epochs on CIFAR100N and CIFAR80N. Experiments are conducted under various noise conditions (“Sym” and “Asym” denote the symmetric and asymmetric label noise, respectively).}
	\label{figure1}
\end{figure*}

\begin{figure*}[h]
	\centering
	\includegraphics[width=0.8\linewidth]{supplementary/supplementary_1.pdf}
	\caption{The test naccuracy (\%) \vs epochs on CIFAR100N and CIFAR80N. Experiments are conducted under various noise conditions (“Sym” and “Asym” denote the symmetric and asymmetric label noise, respectively).}
	\label{figure2}
\end{figure*}

\begin{table*}[t]
\renewcommand\tabcolsep{12pt}
\begin{center}
\centering
\begin{tabular}{c|c|cccc}
\toprule
\textbf{$m$ for SCS and SCR} & \textbf{$\alpha$ for the mean-teacher model} & \textbf{results}\\
\midrule
0.85  &0.85 &57.82\\
0.90  &0.85 &57.92\\
0.95 &0.85 &59.41\\
0.99  &0.85 &62.63\\
0.999  &0.85 &61.60\\
\bottomrule
\end{tabular}
\end{center}
\caption{Effects of hyper-parameter $m$ for SCS and SCR in test accuracy (\%) on CIFAR100N (noise rate and noise type are 0.5 and symmetric, respectively).}
\label{table:2}
\end{table*}

\begin{table*}[t]
\renewcommand\tabcolsep{12pt}
\begin{center}
\centering
\begin{tabular}{c|c|cccc}
\toprule
\textbf{$m$ for SCS and SCR} & \textbf{$\alpha$ for the mean-teacher model} & \textbf{results}\\
\midrule
0.85 &0.90 &57.73\\
0.85 &0.95 &58.04\\
0.85 &0.99 &57.46\\
0.85 &0.999 &57.86\\
\bottomrule
\end{tabular}
\end{center}
\caption{Effects of hyper-parameter $\alpha$ for the mean-teacher model on CIFAR100N (noise rate and noise type are 0.5 and symmetric, respectively).}
\label{table:3}
\end{table*}

\subsection{ Truncated Normal Distribution in SCR. }
Considering the confidence of the corrected label of noisy samples, we propose a dynamic truncated normal distribution as a sample re-weighting function to mitigate the biased label correction.
Samples with higher correction confidence are less likely to be incorrectly labeled compared to those with lower confidence, thus being assigned larger weights.
We further perform an ablation study to explore the effectiveness of re-weighting corrected labels based on their confidence.
As shown in Table~\ref{table:1}, re-weighting corrected labels improves the performance by 0.68\% compared to the without-reweighting case.

In our proposed SCR, we employ a dynamic truncated normal distribution, whose mean and variance values are $\mu_t$ and $\sigma_t$ at the $t$-th epoch, to assign weights for different samples.
Moreover, to enable class-balanced re-weighting and promote training stability, we propose to estimate $\mu_t(c)$ and $\sigma_t^2(c)$ for each class $c$ based on their historical estimations using EMA.
Table~\ref{table:1} shows that employing SCR with EMA achieves an additional 0.35\% performance gain compared to SCR without EMA.

\subsection{Effect of Hyper-parameters in EMA.}
Considering that the model is inevitable to fit some noisy samples in the later stage of training, we resort to the exponential moving average (EMA) to achieve more reliable sample selection, label correction, and sample re-weighting.
EMA introduces the model's results in the historical iteration to increase the stability of the training process.

To dynamically reflect the learning performance changes of the model, we use EMA to update our global and local thresholds in SCS dynamically and the truncated normal distribution parameters in SCR.
Besides, our mean-teacher model ($\theta^*$) is also updated in an EMA manner.
The exponential moving average strategy requires a factor (\ie, $m$ for SCS and SCR, $\alpha$ for the mean-teacher model) to balance the weight of past and current results, thus ensuring the stability of the training process.
As shown in Table~\ref{table:1}, employing SCS with EMA and employing SCR with EMA both achieve nontrivial performance gains.

Table~\ref{table:2} shows the test accuracy of our SED under different $m$ for SCS and SCR with a fixed $\alpha$ (\ie, 0.85) on CIFAR100N-Sym50\%.
It can be observed that the best performance is achieved when $m = 0.99$.
Table~\ref{table:3} further shows the test accuracy of our SED under different $\alpha$ for the mean-teacher model with a fixed $m$ (\ie, 0.85) on CIFAR100N-Sym50\%.
As we observe from Table~\ref{table:3}, the varying alpha values have only a minor effect on the performance of our SED. The best performance is obtained when $\alpha=0.95$.

\section{ Extend Experiment Results }
Due to the memorization effect \cite{memorization-effect} (\ie, models tend to fit clean and simple samples first and then gradually memorize noisy ones), the network optimization based on the cross-entropy loss usually leads to an ill-suited solution.
During our experiments, CE refers to the conventional training baseline that utilizes the entire noisy dataset with the cross-entropy loss.
As shown in Fig. (5) (c), the test accuracy of CE first increases to a certain level and then decreases due to overfitting.
Fig. (5) (c) has already shown that robust methods \cite{dividemix, Small_Loss} for noisy labels like our SED are able to mitigate the impact of overfitting noisy samples in the later stage of training.
The test accuracy of these robust methods increases monotonously as the training continues.

Fig.~\ref{figure2} further verifies the robustness of our SED as the training processes. 
Fig.~\ref{figure2} records the test accuracy of the training process under different noise conditions (\ie, “Sym” and “Asym” denote the symmetric and asymmetric label noise, respectively. 20\%, 40\%, and 80\% denote the noise rate.) on CIFAR100N and CIFAR80N.
It can be observed that the test accuracy continues to grow under various noise conditions.
This explicitly proves that our method can prevent the model from overfitting the noisy samples, thereby enhancing its robustness.

To further verify the effectiveness of our proposed SED in practical scenarios, we conduct experiments on three real-world noisy datasets (\ie, Web-Aircraft, Web-Car, and Web-Bird \cite{webfg}), whose training images are crawled from web image search engines.
Fig.~\ref{figure3}, Fig.~\ref{figure4}, and Fig.~\ref{figure5} show the comparison of the sample selection results between our SED and two state-of-the-art (SOTA) methods (\ie, JoCoR \cite{JoCoR} and Co-LDL \cite{Co-LDL}) on these three real-world datasets.
These sample selection methods primarily seek to split samples into two subsets: a noisy subset and a clean subset.
The samples marked with purple represent the incorrectly selected samples.
The samples marked with green are the correctly selected samples.
If the sample selection is correctly performed in all three methods, the corresponding mark is displayed in black.
Fig.~\ref{figure3}, Fig.~\ref{figure4}, and Fig.~\ref{figure5} visually illustrate the excellent performance of our SED.

\begin{figure*}[t]
	\centering
	\includegraphics[width=\linewidth]{supplementary/supplementary_5.pdf}
 \vspace{-0.5cm}
	\caption{The comparison of sample selection results with SOTA approaches on Web-Aircraft. Each row shows seven samples that are randomly selected from the same fine-grained category. Images in the red box indicate noisy samples.}
	\label{figure3}
\end{figure*}

\begin{figure*}[t]
	\centering
	\includegraphics[width=\linewidth]{supplementary/supplementary_3.pdf}
 \vspace{-0.6cm}
	\caption{The comparison of sample selection results with SOTA approaches on Web-Bird. Each row shows seven samples that are randomly selected from the same fine-grained category. Images in the red box indicate noisy samples.}
	\label{figure4}
\end{figure*}
\begin{figure*}[t]
	\centering
	\includegraphics[width=\linewidth]{supplementary/supplementary_4.pdf}
 \vspace{-0.8cm}
	\caption{The comparison of sample selection results with SOTA approaches on Web-car. Each row shows seven samples that are randomly selected from the same fine-grained category. Images in the red box indicate noisy samples.}
	\label{figure5}
\end{figure*}


\bibliographystyle{splncs04}
\bibliography{egbib}